# modelDNA: Calibrated Lineage Verification and Merge Decomposition from Sampled Weight Fingerprints

**Muhammad Awais Bin Adil · Saad Aamir**
Independent · binadilawais@gmail.com · saadaamir473@gmail.com

Code: https://github.com/AwaisAdilKhokhar/modelDNA · Live scanner: https://huggingface.co/spaces/AwaisAdilKhokhar/modelDNA · Data: https://huggingface.co/datasets/AwaisAdilKhokhar/modeldna-atlas

---

## Abstract

The lineage graph of open-weight language models is self-reported: Hugging Face's `base_model` metadata field is optional and unverified, and over 60% of Hub models document no parentage at all. Methods for detecting lineage from weights exist in the research literature, but each ships as paper code tied to one signal and one experiment; when a provenance dispute breaks, the analysis is redone by hand. This report describes modelDNA, a tool that fingerprints a model from roughly 100-300 MB of ranged HTTP reads (instead of a full 15 GB download for a 7B model), compares the fingerprint against a reference database of foundation models across four published signal families, and returns one of eight verdict classes with a calibrated probability, preferring honest abstention to confident error. On a benchmark of 15 real Hub models with org-documented parentage, judged against 8 candidate bases (13 positives, 107 hard negatives), the system achieves AUROC 1.0, zero false positives at its reporting threshold, and 13/13 correct top-1 parent attribution. The report's second contribution is merge decomposition. Every mainstream weight-merging method is (near-)linear per tensor, and fingerprint sample positions are deterministic functions of tensor identity, so a merged model's fingerprint is the same linear combination of its parents' fingerprints. Mixture weights can therefore be recovered from fingerprints alone by sum-to-one constrained least squares, which is algebraically a well-conditioned regression in task-vector space. Against merges with published mergekit configurations as ground truth, the method recovers a slerp merge's layer-interpolation curves at r = 0.999 and a dare_ties merge's mixture weights to within 0.011 of the published values, without downloading any weights beyond the fingerprints. All fingerprints, benchmarks, and the inferred lineage graph of 55 models are public and reproducible offline.

---

## 1. Introduction

Hugging Face hosts nearly three million model repositories (2.9M as of July 2026), and the graph of who fine-tuned whom is almost entirely on the honor system. The `base_model` metadata field is optional, unverified, and frequently absent: Horwitz et al. [5], building their Model Atlas, measured that more than 60% of Hub models carry no documented parentage, and had to manually inspect the largest connected components of their own dataset just to fill in missing values. Hugging Face itself ran a volunteer sprint asking the community to open metadata PRs by hand.

This matters beyond bookkeeping. Three well-documented incidents shaped this project. In July 2025, an anonymous group published an analysis showing a 0.927 correlation between the per-layer attention-projection standard-deviation curves of Huawei's Pangu Pro MoE and Alibaba's Qwen2.5-14B, against an unrelated-pair background of roughly 0.3-0.7 [1]; Huawei denied derivation, the analysis repo was taken down, and the dispute ran in the international press with no neutral party able to re-run the numbers. In September 2024, Reflection-70B was marketed as a breakthrough fine-tune until community weight-diffing showed the claims didn't hold up. In June 2024, the Llama3-V project was shown to be substantially copied from MiniCPM-Llama3-V 2.5. In each case the investigation was ad hoc: someone with the right skills happened to spend a weekend on it.

The detection methods themselves are not the bottleneck. At least ten papers from 2024-2026 demonstrate lineage signals that survive fine-tuning, continued pretraining, merging, pruning, quantization, and in some cases deliberate re-parameterization (Section 2). What has been missing is the unglamorous part: a maintained tool that takes a repo id and returns a verdict, with a reference database, calibrated probabilities, background distributions, and an explicit statement of what the evidence does and does not show. modelDNA is that tool: `pip install modeldna`, then `modeldna scan org/model`.

The design is shaped by one asymmetry: the worst failure mode of a lineage tool is not a missed detection but a false accusation. The conservative thresholds, the abstention band, the zero-false-positive ship gate, and the background distributions printed next to every score all follow from that asymmetry. They trade recall for precision on purpose.

This report makes four contributions:

1. **A fingerprint design** that captures four published signal families from roughly 100-300 MB of byte-range reads per 7B model, with sample positions that are pure functions of (seed, canonical role, layer, tensor size). This property makes fingerprints of different models directly comparable, and it later turns out to enable merge decomposition (Section 4).
2. **A calibrated, abstention-first verdict engine** that maps pairwise evidence to one of eight classes with a probability, and imputes missing signals with unrelated-pair background means so absence of evidence never counts as evidence (Section 5).
3. **Merge decomposition from fingerprints alone**: recovering the mixture weights of a merged model by constrained least squares on fingerprint samples, validated against merges whose mergekit configurations are public (Sections 6, 7.3). To our knowledge this is the first demonstration that merge recipes can be read back out of sampled fingerprints without access to full weights.
4. **LineageBench**, a benchmark of real Hub models whose parentage is corroborated by the publishing organization's own documentation rather than by the metadata tag the tool exists to audit, plus committed fingerprint caches that make every number in this report reproducible offline (Section 7.2).

## 2. Related work

**Intrinsic weight-space fingerprints.** HonestAGI [1] formalized lineage detection as a binary decision (derived through continued training versus trained independently) using the per-layer standard deviations of attention projection matrices, normalized across depth and compared by correlation; the signature survives continued pretraining, and the paper's background statistics (unrelated pairs at r of roughly 0.3-0.7, derived pairs above 0.9) anchor modelDNA's primary signal. HuRef [2] established that base-model parameter *directions* barely move under SFT and RLHF, and defined the PCS comparator (cosine over flattened weights) that later work uses as a baseline; PCS is the cheapest decisive test for the non-adversarial majority of Hub models. *Ghost in the Transformer* [3] showed that singular-value spectra detect reuse while being inherently invariant to orthogonal re-parameterizations, which direct-similarity signals are not. modelDNA productizes all three families plus 1-D vector comparisons, and treats them as an ensemble with per-signal reporting rather than picking a winner.

**Population-scale lineage recovery.** MoTHer [4] and Model Atlas [5] reconstruct directed family trees over whole repository populations from weight distances and structural priors. Both papers note the same two open failure modes, undocumented merges and distillation. Merges are the case Section 6 addresses. Distillation is invisible to weight forensics by construction and is stated as out of scope everywhere modelDNA reports a verdict.

**Behavioral and gradient-based methods.** REEF [7] compares activation similarity on shared inputs and has the strongest published robustness envelope, at the cost of GPU forward passes on both models; TensorGuard [8] fingerprints gradient responses to input perturbations; SeedPrints [9] shows initialization-seed biases persist through training and provides calibrated significance tests; PhyloLM [10] builds phylogenies from outputs alone. These are complementary deeper tiers, not competitors. modelDNA's scope in this report is deliberately the static, CPU-only, weights-as-artifact regime, because that is the regime in which a scan can run in two minutes on a laptop against any Hub repo.

**Injected fingerprints and watermarks.** A separate literature embeds owner-chosen keys into models before release (instructional fingerprinting, Chain & Hash, and successors; see the SoK in [11]). modelDNA is strictly passive forensics on models the operator does not own, so these methods are out of scope, though the taxonomy (injected versus intrinsic) is the right frame for where this tool sits.

**Adversarial limits.** Evasion attacks defeat fingerprints previously believed robust [6]. The consequences adopted here: never claim adversarial completeness; stack independent signal families so evasion cost compounds; and report which signals matched, so that "low direct similarity, high invariant similarity" is surfaced as an anomaly pattern that is itself informative (Section 8).

Relative to all of the above, this report's claims are deliberately modest on signal novelty (the four families are from the literature) and concrete on three things the literature does not provide: the composition of calibration, abstention, background reporting, and one command; the sampling design that makes fingerprints of different models element-aligned; and the demonstration that this alignment is sufficient to decompose merges.

## 3. Problem setting

Given a *suspect* model on the Hub (or a local directory) and a reference database of candidate ancestors, decide which of the following the weight evidence supports, and with what confidence: the suspect is a copy, a quantized copy, a fine-tune, or a heavier continuation of some candidate; a merge of several; related at family level only; or unrelated to everything indexed. Two constraints make this different from the pairwise-detection setting the papers evaluate:

- **The tool must scale down, not up.** A verdict should not require a GPU, a full download, or the suspect's cooperation. Everything runs CPU-only from ranged HTTP reads.

- **The output is public-facing.** A score without context convinces nobody (a lesson from the Pangu dispute, where a bare 0.927 was argued about for days until it was shown against the unrelated-pair background). Every number modelDNA reports ships with the background distribution measured on its own database, and the phrasing is always “weights are statistically consistent with derivation from X”, never an accusation.

## 4. Fingerprints

### 4.1 Reading a model without downloading it

The safetensors format stores a JSON header (tensor names, dtypes, shapes, byte offsets) at the start of each shard, and the Hub serves byte ranges over HTTP. modelDNA reads the headers (kilobytes), plans every byte range the extraction will need, and fetches the plan concurrently with ranges coalesced per shard. A 7B model fingerprints in about two minutes over an ordinary connection, reading 100-300 MB instead of roughly 15 GB. The planning step is possible for the same reason the fingerprints are comparable at all: sample positions are computed from tensor identity, not from tensor content, so the complete read plan is known before any data moves.

GGUF, the llama.cpp/ollama distribution format and a large fraction of what is actually downloaded, is read the same way, by *dequantize-and-sample*. Every ggml quantization type packs a fixed number of elements into a fixed number of bytes per block, so a sampled element range maps to a block-aligned byte range exactly as it does for safetensors: fetch the covering blocks, dequantize with llama.cpp’s reference kernels, trim to the requested positions. One conversion quirk must be undone for positions to align: the HF→GGUF converter permutes attention q/k rows for llama-family models. With that permutation reversed, a quantized copy compares against its full-precision parent with only quantization noise in between, instead of abstaining. Validated live: a Q4_K_M of Mistral-7B-v0.1 fingerprints from 149 MB of a 4.4 GB file and scores PCS 0.998 against its bf16 parent versus 0.003 against an unrelated same-shape model.

### 4.2 Four signal families

Tensor names are first mapped to canonical roles (`attn.q`, `attn.k`, `mlp.down`, `norm.input`, …) so that models with different naming schemes are comparable. A fingerprint then bundles:

| | signal | what is stored | follows | survives |
|---|---|---|---|---|
| F1 | attention σ-curves | per-layer std of each attention/MLP projection | [1] | SFT, continued pretraining, LoRA-merge |
| F2 | 1-D vectors | norm gains and biases: per-layer L2 norms + subsamples, read in full | [1] | heavy training |
| F3 | PCS samples | seeded element samples of the 2-D projections | [2] | SFT; the cheapest decisive test |
| F4 | spectra | top-32 singular values per sampled layer | [3] | permutation / rotation re-parameterization |

The sampling scheme is the load-bearing design decision. For each 2-D tensor, a seed is derived by hashing (global seed, canonical role, layer index), never the raw tensor name, and drives 32 pseudo-random contiguous blocks of 16,384 elements each. Because the positions depend only on (seed, role, layer, tensor size), any two models with matching shapes are sampled at *identical element positions*. A cosine between two models' PCS samples is therefore an unbiased estimate of the cosine between the full flattened tensors, from about half a million elements per role instead of billions. σ-curves are estimated from the same samples (the standard error of a std estimate over 512k elements is negligible at these matrix sizes). F2 vectors are small enough to read exactly. F4 spectra are estimated in fast mode from a seeded 192-row band of every fourth layer's projection via randomized subspace iteration [12], and exactly in full mode.

The sampling constants are frozen and versioned: changing any of them silently invalidates every stored fingerprint, so they are treated as part of the schema.

A fingerprint serializes to roughly 1.5 MB gzipped. The reference database covers 37 foundation bases spanning Llama, Qwen, Mistral, Gemma, Phi, DeepSeek, Yi, Falcon, OLMo and others, plus deliberate hard negatives such as OpenLLaMA (Llama shapes, independent training). It ships as a 57 MB archive that installs in seconds, so a scan only ever fetches the suspect. No weights are stored or redistributed at any point, which also keeps the database license-clean.

## 5. From evidence to verdict

### 5.1 Pairwise evidence

Comparing two fingerprints yields per-role correlations and cosines for each signal family, plus structural facts (layer counts, shape compatibility, tensor-inventory and tokenizer hashes). Structure acts as a hard filter: a layer-count mismatch rules out per-layer signals and routes to the depth-surgery limitation path rather than producing a low score that could be mistaken for evidence of independence.

### 5.2 Calibration

A logistic model over the four aggregate features (mean σ-curve correlation, mean vector cosine, mean PCS cosine, mean spectral correlation) produces the probability. Two details matter more than the model class, which is deliberately boring:

- **Missing features are imputed with unrelated-pair background means.** A signal that could not be computed (PCS skipped because shapes differ, say) pushes the score toward the background, so absence of a signal is never evidence in either direction.
- **Coefficients are constrained non-negative.** Every feature is oriented "higher = more similar", so a negative weight can only be an overfitting artifact, and one an adversary could exploit by inflating the corresponding signal. The constraint costs nothing on the benchmarks and closes that door.

The shipped coefficients are a bootstrap fit on a synthetic benchmark (Section 7.1); the probability *rankings* on real models are validated in Section 7.2, and the honest caveat about probability *magnitudes* is in Section 8.

### 5.3 Verdict classes and abstention

Every scan resolves to exactly one of eight classes: `EXACT_COPY`, `QUANTIZED_COPY`, `FINE_TUNE`, `SAME_LINEAGE`, `LIKELY_MERGE`, `SAME_FAMILY_UNRESOLVED`, `NO_MATCH`, `INSUFFICIENT`. The positive threshold is $p \geq 0.90$; candidates in the 0.50-0.90 band produce `SAME_FAMILY_UNRESOLVED` ("suggestive, not conclusive") rather than a call. Two positives in different families produce `LIKELY_MERGE` with a pointer to the decomposition tool. Two positives in the same family within $\Delta p \leq 0.03$ of each other are declared a tie the data cannot break, which is the honest answer when, for example, a fine-tune matches both Mistral-7B-v0.1 and v0.3, two bases that are themselves closely related. Subclassing among positives (copy / quantized / fine-tune / heavier lineage) is driven by PCS magnitude and structural hashes.

The claimed-versus-detected consistency flag fires only when the repo makes an explicit lineage claim that the weight evidence contradicts at high confidence. A missing `base_model` tag is never "inconsistent"; it is just missing.

## 6. Merge decomposition

Merges are the hard case flagged as open by the graph-recovery literature [4, 5], and they are common: model merging is a large, active subculture with its own tooling (mergekit [13]) and leaderboard presence. The question that community actually asks is not "is this a merge?" but "what is *in* this merge, and in what proportions?" This section shows that fingerprints as designed in Section 4 already contain the answer.

### 6.1 Merges are linear; sampling commutes with linearity

Every mainstream mergekit method produces, per tensor, a weight matrix that is a linear or near-linear combination of the parents' corresponding tensors. Linear interpolation is linear by definition. Task arithmetic, $W = B + \Sigma_i \lambda_i(W_i - B)$ over base B, is itself a sum-to-one linear combination once the base is included as a term with weight $1 - \Sigma_i\lambda_i$. Slerp is not linear in general, but merge parents are fine-tunes of a shared base with pairwise cosines of 0.99 and above; at those angles slerp coincides with linear interpolation up to a correction quadratic in the (tiny) angle. TIES and DARE apply elementwise sign-election and random sparsification to the task vectors, which breaks exact linearity per element but preserves it in expectation; the deviation appears as fit residual rather than as bias in the fitted weights, which Section 7.3 confirms empirically.

Because PCS sample positions are functions of tensor identity only, sampling commutes with any per-tensor linear combination: the merged model's PCS vector is the same mixture of the parents' PCS vectors,

$y \approx \Sigma_i \alpha_i x_i$,

where y and the $x_i$ are the fingerprint sample vectors of the target and candidates. The mixture weights of a 7B merge are therefore recoverable from a least-squares fit over about 500k sampled elements that the fingerprints already contain. No weights are downloaded; parents already in the reference database are not even re-fetched.

### 6.2 The sum-to-one constraint is the whole trick

The naive regression fails in practice, and understanding why dictates the fix. Merge parents share a base, so their raw sample vectors are nearly collinear: cosines of 0.99 and up, every column dominated by the same base direction. Unconstrained least squares on such a design matrix is ill-conditioned, and its output is noise.

Constraining the weights to sum to one changes the problem's geometry. Writing the pivot column's weight as one minus the sum of the others reduces the constrained problem to unconstrained least squares on differences:

$y - x_0 \approx \Sigma_i \beta_i (x_i - x_0)$,

where $x_0$ is the pivot (the shared base, when provided) and the sum runs over the remaining candidates. The regressors $x_i - x_0$ are sampled *task vectors*, the fine-tuning deltas, and fine-tune directions from different training runs are nearly orthogonal, so the reduced system is well-conditioned. The constraint is not a modeling assumption bolted on for convenience; it is what moves the regression from the degenerate raw-weight space into the well-behaved task-vector space. When candidates' task vectors *are* correlated (fine-tunes of a fine-tune, sibling models trained on overlapping data), the fit detects it and says so: pairwise task-vector correlations above 0.95 emit an explicit "their split is poorly identified" warning rather than a silently arbitrary answer.

### 6.3 The evidence statistic, and what the tool refuses to conclude

Reconstruction cosine is too forgiving to serve as merge evidence, because the shared-base direction guarantees a high cosine for almost any mixture. The statistic that actually separates hypotheses is the residual shrink versus the best single candidate:

gain = 1 − RSS(mixture) / RSS(best single parent).

An exact merge scores near 1, and a merge followed by fine-tuning scores around 0.4. The case that sets the threshold is the impostor: a model that is not a merge at all but a sibling fine-tune of one candidate scores around 0.25, because adding an irrelevant second parent always buys a little residual by chance in a near-collinear family. The MERGE call requires gain ≥ 0.30, above the impostor regime; below it the tool reports the fitted weights but labels the summary AMBIGUOUS or SINGLE_PARENT. Lowering that threshold to "fix" a known-merge case would trade directly against false MERGE calls on ordinary fine-tunes, and Section 7.3 includes a real case where the honest answer is exactly this hedge.

Beyond the pooled weights, the decomposition refits per role and per layer (within a role, every layer contributes an equal-length sample chunk, so layers can be split back out). Per-layer fits recover depth-varying mixtures, the interpolation curves of slerp and gradient merges, and this turns out to be the most striking validation result. An independent cross-check fits the same mixture on the 1-D norm/bias vectors (F2), a different signal family entirely; divergence between the two fits is reported as a caution. Negative fitted weights are surfaced with a note (task-vector subtraction is a real mergekit technique, but collinearity warnings should be checked first), and depth-changing merges (passthrough, frankenmerges) are rejected up front rather than decomposed incorrectly.

## 7. Evaluation

Three layers, in increasing order of realism. All fingerprint caches are committed to the repository; every number below reproduces offline with `--no-fetch`.

### 7.1 Synthetic gates

A generated population of independent bases, fine-tunes at varying strengths, heavy continuations, fp16 recasts, and independent same-architecture controls runs in CI on every change and enforces the ship gates: AUROC ≥ 0.99, TPR ≥ 0.95 at FPR ≤ 1%, and zero false positives on independent same-architecture controls at the reporting threshold. The last gate is the defamation-risk surface and is non-negotiable. Current results: AUROC 1.0, zero false positives, maximum calibration gap 0.045. Synthetic tests are a floor, not a claim; they exist to make regressions loud.

### 7.2 LineageBench: real models, org-documented ground truth

Ground truth for a lineage benchmark cannot come from the `base_model` tag, because auditing that tag is the tool's purpose. LineageBench instead uses 15 suspect models across 6 families whose parentage is corroborated by the publishing organization's own documentation: technical reports, official release notes, model cards from the org that trained the parent. The suspects cover the derivation regimes a scanner meets in the wild: community fine-tunes (Zephyr-7B-β, OpenHermes-2.5, Nous-Hermes, Dolphin-2.9), continued pretrains (CodeLlama-7B, Qwen2.5-Coder-7B-Instruct), official instruct releases (Llama-3, Qwen2.5, Mistral, Gemma), a merge chain (AlphaMonarch-7B), a depth-upscale (SOLAR-10.7B), and a GPTQ quantized copy. Each suspect is judged by the actual verdict engine against all 8 candidate parents: the true parent must win, not merely score well in isolation.

Thirteen suspects form the positive pool. The negative pool contains 107 hard negatives: every positive suspect scored against each cross-family wrong parent, plus all cross-family parent-versus-parent pairs. (Same-family wrong parents, such as Zephyr against Mistral-v0.3 rather than v0.1, are excluded from the negative pool because they genuinely share lineage; scoring them as negatives would punish the tool for detecting something true.) The depth-upscale and quantized-copy cases are excluded from the scored pool by design and evaluated separately against their documented expected behavior, which is abstention.

Results:

| metric | value |
|---|---|
| AUROC (13 positives vs 107 hard negatives) | 1.0 |
| TPR at 1% FPR | 1.0 |
| false positives at the $p \geq 0.9$ reporting threshold | 0 / 107 |
| weakest positive | 0.9426 (gemma-2b-it) |
| strongest negative | 0.6596 |
| top-1 parent attribution | 13 / 13 |

The weakest positive by derivation kind: fine-tunes 0.998, merge-chain 0.9984, continued pretrains 0.9796, official instructs 0.9426. The separation between the weakest positive and the strongest negative is clean, with the entire abstention band (0.50-0.90) empty on this data.

The two limitation cases behaved exactly as documented. SOLAR-10.7B, a depth-upscale from 32 to 48 layers, produced an honest `NO_MATCH` with the layer-count mismatch reported; the tool does not attempt a depth-alignment search yet, and it says so rather than guessing. The GPTQ copy abstained on the packed int4 tensors while still ranking the true parent first on the signals it could compute.

Two caveats, reported rather than hidden. First, AlphaMonarch-7B, the merge-chain case, resolved to the correct family as `SAME_FAMILY_UNRESOLVED` rather than `LIKELY_MERGE`: all its ancestors are Mistral fine-tunes, so no cross-family split exists for the scan heuristic to notice, and the scan-level merge flag is family-based in v0.1. (Section 7.3 shows that the decomposition tool, given candidates, resolves this model completely.) Second, Nous-Hermes-llama-2-7b was attributed to the correct parent but classed `QUANTIZED_COPY` rather than `FINE_TUNE`: its fine-tune moved the weights so little that its PCS cosine against Llama-2-7b (0.9998) crosses the dtype-recast rule's 0.999 threshold, and because the repos ship in different dtypes (bfloat16 versus the parent's float16) the recast rule fires before the fine-tune rule. The parent, probability, and ranking are all right, so the top-line metrics are unaffected, but the boundary between "recast of the same values" and "very light fine-tune" needs a delta-based discriminator rather than a cosine threshold. We would rather record that here than have a reader discover it.

A scoping note: at 13 positives and 107 negatives this benchmark is deliberately smaller than the 300-pair target the project's own planning document sets. Every ground-truth label carries a citation to org documentation, but the AUROC of 1.0 should still be read as "no errors at this scale", not as a claim that errors are impossible. Growing the benchmark is mechanical (the fingerprint pipeline is the expensive part, and it is now fast) and is the highest-priority follow-up.

### 7.3 Merge decomposition against published recipes

The strongest available ground truth for merge decomposition is a merge whose author published the exact mergekit configuration. Three cases, all decomposed from fingerprints alone:

**NeuralPipe-7B-slerp** (slerp of two Mistral fine-tunes, published interpolation anchors `[0, 0.5, 0.3, 0.7, 1]` for attention and `[1, 0.5, 0.7, 0.3, 0]` for MLP, each interpolated evenly across the 32 layers). The pooled fit reports MERGE with weights 0.644 / 0.356 and reconstruction cosine 0.99996; the mixture removes 78% of the residual the best single parent leaves. The result worth the reader's attention is the per-layer fit, shown in Figure 1: the recovered layer-by-layer mixture weights trace the model card's *opposite* attention and MLP interpolation curves at r = 0.999 (attention, max error 0.048) and r = 0.970 (MLP). The MLP fit's only visible deviation sits at layer 0, where the parents' task-vector norm collapses to near zero; at that layer the two parents are nearly identical and the weight is mathematically unidentifiable, not misestimated. The independent F2 cross-check diverges from the per-role curves exactly as it should: the published config merges the norm vectors at a flat t = 0.5 while the projections follow the curves, and the tool flags the divergence.

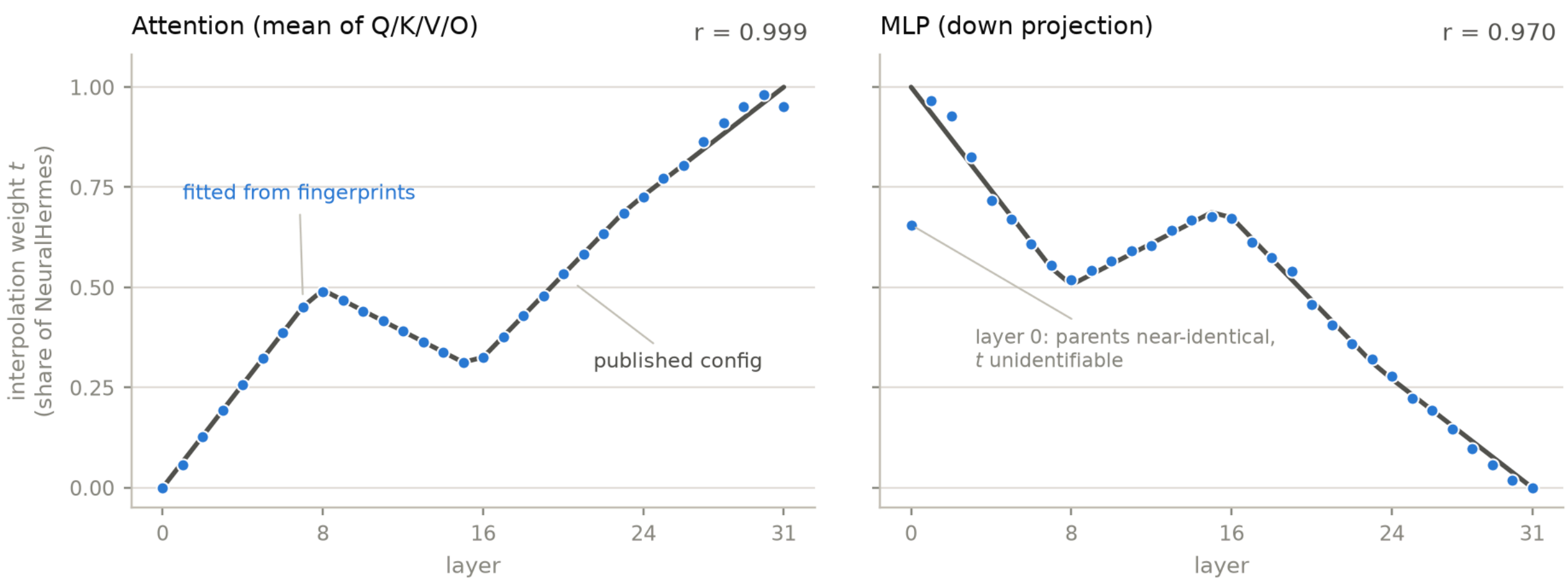

*Figure 1. Reading a merge recipe back out of fingerprints. Each dot is the sum-to-one least-squares weight of one parent (NeuralHermes) fitted on that layer's fingerprint samples; the line is the interpolation curve the merge's author published. Regenerate with* `python paper/make_figures.py`.

**Monarch-7B** (dare_ties over three Beagle-family fine-tunes, published weights 0.36 / 0.34 / 0.30, density 0.6). The fit recovers 0.371 / 0.347 / 0.291 (maximum error 0.011, cross-role spread under 0.01), with the base column correctly near zero (−0.01). The summary, however, is AMBIGUOUS, not MERGE, and this is the honest hedge described in Section 6.3 doing its job: two of the three parents' task vectors are correlated at r = 0.987 (the collinearity warning fires), and DARE sparsification at density 0.6 injects elementwise noise comparable in magnitude to the task vectors themselves, which bounds the achievable residual shrink to 0.219, below the 0.30 MERGE threshold that keeps sibling fine-tunes from being mislabeled as merges. The mixture weights are right; the tool declines to *certify* the merge from residual evidence alone, and says why. A promising discriminator for the next version: DARE residuals are heavy-tailed and elementwise-correlated with the parent deltas, while post-hoc SFT residuals are not.

**AlphaMonarch-7B** (chain closure). Given its true immediate ancestor Monarch-7B plus two decoys, the fit puts α = 1.000 on Monarch, roughly 0 on both decoys, and reports SINGLE_PARENT with essentially zero gain, which is the correct description of a model that is a descendant, not a merge, of the listed candidates. This case also resolves a pitfall discovered during development: with an *incomplete* candidate list (grandparents but not the true parent), the fit spreads weight across deep ancestry (α of 0.44 on a grandparent) and correctly refuses the MERGE label on gain, but the fitted weights are then describing ancestry bleed-through, not a recipe. Decomposition answers "how does this model decompose over *these* candidates", and the candidate list is part of the question.

Live behavior matches the committed benchmarks: the hosted scanner's decompose tab reproduces the NeuralPipe fit exactly (0.644 / 0.356) on Space hardware in about 200 seconds, fetching only fingerprint-sized reads for the models not already in the reference DB.

## 8. Limitations

**Statistical consistency, not proof.** Every output is "weights are statistically consistent with derivation from X" against a measured background. That is the strongest claim the method supports, and the report format never exceeds it.

**Distillation is invisible**, to this tool and to every weight-space method, by construction: a distilled student shares no weight linkage with its teacher. Behavioral tiers (REEF-style activation comparison, output phylogenies) are the roadmap answer and will be labeled experimental when they land.

**Merge decomposition needs a candidate list.** A scan flags `LIKELY_MERGE` only on cross-family splits, and, as AlphaMonarch shows, merges within one family present as family-level matches at scan time. Decomposition then requires naming candidates, and an incomplete candidate list yields weights that describe ancestry rather than recipe. Auto-suggesting candidates from the scan's positive set is planned but not built. Depth-changing merges are rejected, not decomposed.

**Adversarial robustness is bounded, and stated.** Deliberate permutation or rotation re-parameterization defeats the direct-similarity signals (F1-F3); the spectral family (F4) is the in-scope counter, and the anomaly pattern "low direct similarity, high invariant similarity" is surfaced as itself informative. Published evasion attacks [6] defeat fingerprints previously believed robust; a retraining-scale adversary is out of scope and every report footer says so.

**Probability magnitudes are less validated than rankings.** The calibrator is a bootstrap fit on synthetic pairs; on LineageBench its *rankings* are perfect but 13 positives cannot meaningfully validate calibration curves, and the weakest real positives (official instructs at 0.94) suggest the probabilities run conservative on heavily-tuned derivatives. Refitting on a larger LineageBench is planned; until then, thresholded verdicts and rankings are the trustworthy output, and exact probability values should be read with a grain of salt.

**Coverage.** Safetensors and GGUF are read natively; GGUF quants of any depth are dequantized at the sample positions and compared directly (Section 4.1), with one honest caveat: at Q4 and below, quantization noise and a light fine-tune occupy the same PCS band, so a deep quant of a base classifies as `FINE_TUNE` with an explicit note that it may equally be a quantized copy; the parent attribution itself is unaffected. Still out: repos shipping exclusively PyTorch `.bin` checkpoints (4 of the 41 seed bases fall out for this reason) and GPTQ/AWQ int4 packing, which abstains rather than being unpacked. Both remaining gaps are engineering gaps rather than method limits.

**Scale.** Fifteen benchmark suspects, 55 models in the public graph, 37 indexed bases. Nothing in the method caps this (fingerprinting takes about 2 minutes per model and the pipeline is automated), but the numbers in this report are the numbers, and they come with the error bars of their size.

## 9. Responsible use

The tool's framing assumes its outputs will end up in disputes. Three rules are enforced in the product, not just the documentation. Verdicts are phrased as statistical consistency and carry their background distributions; the consistency flag fires only against explicit claims, never against silence; and thresholds are set so that the shipped configuration has produced zero false positives across both benchmark suites, with the abstention classes absorbing every case where the evidence is real but insufficient. Reproducibility is the other half of accountability: any verdict can be re-derived by anyone with `modeldna compare a b --report`, a pinned database version, and no privileged access. That anyone can re-run the math is exactly what the Pangu dispute lacked.

## 10. Artifacts

Everything in this report is public:

- **Tool**: https://github.com/AwaisAdilKhokhar/modelDNA (`pip install modeldna`, Apache-2.0)
- **Live scanner**: https://huggingface.co/spaces/AwaisAdilKhokhar/modelDNA. Paste a repo id, verdict in about 2 minutes.
- **Atlas**: https://awaisadilkhokhar.github.io/modelDNA/. The inferred family tree of 55 models (475 depth-compatible pairwise comparisons, 116 edges), every edge carrying its evidence, decomposed merges drawn with their fitted mixture weights.
- **Dataset**: https://huggingface.co/datasets/AwaisAdilKhokhar/modeldna-atlas. Fingerprints, pairwise evidence, the edge list, and the pullable reference DB, refreshed weekly by CI.
- **Benchmarks**: `benchmarks/` in the repository, with ground-truth manifests carrying per-pair citations, committed fingerprint caches, and harnesses that reproduce every number here offline (`--no-fetch`).
- **LineageBench**: the real-model benchmark of Section 7.2 as a named public dataset, https://huggingface.co/datasets/AwaisAdilKhokhar/lineagebench, with a standalone write-up at `paper/lineagebench.md`.

---

## Appendix A: Reproducing every number in this report

```
git clone https://github.com/AwaisAdilKhokhar/modelDNA && cd modelDNA
pip install -e .

# Section 7.1: synthetic gates
python benchmarks/synthetic_bench.py

# Section 7.2: LineageBench, offline from the committed fingerprint cache
python benchmarks/real_lineagebench.py --no-fetch

# Section 7.3: merge decomposition vs published mergekit configs
python benchmarks/merge_decompose_bench.py

# The Atlas and dataset
python scripts/build_atlas.py
python scripts/export_dataset.py
```

Fingerprint caches for all benchmark models are committed under `benchmarks/`, so none of the above touches the network. Ground-truth labels, each with a citation to the publishing organization's documentation, are in `benchmarks/lineagebench_pairs.json`.